\documentclass[10pt,twocolumn,letterpaper]{article}

\usepackage[pagenumbers]{cvpr} 

\newcommand{\rowlightgray}{\rowcolor{gray!15}}

\usepackage{xcolor}
\definecolor{darkgreen}{RGB}{0, 150, 0}
\newcommand{\greensub}[1]{\textsubscript{\footnotesize\textcolor{darkgreen}{#1}}}

\definecolor{darkred}{RGB}{180, 0, 0}

\usepackage{xcolor}
\usepackage{tcolorbox}
\usepackage{fancyvrb}
\tcbuselibrary{skins, breakable}
\newtcolorbox{promptbox}[1]{
    colback=gray!10,           
    colframe=black,            
    arc=3mm,                   
    boxrule=1pt,               
    left=4mm, right=4mm, top=3mm, bottom=3mm,  
    fonttitle=\bfseries,       
    title={#1},                
    breakable                  
}

\tcbuselibrary{listings}
\tcbset{
    listing style/.style={
        listing only,
        colback=white,
        colframe=black,
        boxrule=0.5pt,
        sharp corners,
        left=6pt, right=6pt, top=6pt, bottom=6pt,
        fonttitle=\bfseries,
        listing options={
            basicstyle=\ttfamily\small,
            breaklines=true,
            columns=fixed,
            keepspaces=true,
            linewidth=\textwidth
        }
    }
}

\renewcommand{\paragraph}[1]{\vspace{.5em}\noindent\textbf{#1}}

\definecolor{cvprblue}{rgb}{0.21,0.49,0.74}
\usepackage[pagebackref,breaklinks,colorlinks,allcolors=cvprblue]{hyperref}

\usepackage{graphicx}
\usepackage{float}
\usepackage{multirow}
\usepackage{multicol}
\usepackage{makecell}
\usepackage{pifont}
\usepackage{bbm}
\usepackage[table]{xcolor}
\usepackage{pifont}
\usepackage{url}

\def\paperID{} 
\def\confName{CVPR}
\def\confYear{2026}

\title{Aligning What Vision-Language Models See and Perceive with Adaptive Information Flow}

\author{Chengxin Liu$^1$ \quad Wonseok Choi$^2$ \quad Chenshuang Zhang$^1$ \quad Tae-Hyun Oh$^1$\\
		$^1$KAIST \quad
		$^2$POSTECH\\
		{\tt\small \{cxliu, taehyun.oh\}@kaist.ac.kr}
	}

\begin{document}
\maketitle
\begin{abstract}

Vision-Language Models (VLMs) have demonstrated strong capability in a wide range of tasks such as visual recognition, document parsing, and visual grounding. Nevertheless, recent work shows that while VLMs often manage to capture the correct image region corresponding to the question, they do not necessarily produce the correct answers. 
In this work, we demonstrate that this misalignment could be attributed to suboptimal information flow within VLMs, where text tokens distribute too much attention to irrelevant visual tokens, leading to incorrect answers.
Based on the observation, we show that modulating the information flow during inference can improve the perception capability of VLMs.
The idea is that text tokens should only be associated with important visual tokens during decoding, eliminating the interference of irrelevant regions. To achieve this, we propose a token dynamics-based method to determine the importance of visual tokens, 
where visual tokens that exhibit distinct activation patterns during different decoding stages are viewed as important. We apply our approach to representative open-source VLMs and evaluate on various datasets, including visual question answering, visual grounding and counting, optical character recognition, and object hallucination. The results show that our approach significantly improves the performance of baselines. 
Project page: \href{https://cxliu0.github.io/AIF/}{https://cxliu0.github.io/AIF/}.

\end{abstract}    
\section{Introduction}
\label{sec:intro}

The community has witnessed significant progress in VLMs over the past few years. 
With continuous innovations in model architecture~\cite{dai2023instructblip,liu2024llava,chen2024internvl,bai2025qwen25} and model training~\cite{guo2025seed,internvl3}, VLMs can now function as a general-purpose assistant, showcasing strong multimodal understanding and reasoning capabilities in a wide range of tasks, such as visual recognition~\cite{realworldqa,mmstar2024,tong2024mmvp,mmbench}, object grounding~\cite{yu2016refcoco,refcoco2014,nagaraja2016refcoco}, and document parsing~\cite{chartqa2022,seedbench,textvqa}. 

Despite being powerful, it has been reported that VLMs have 
misalignment between seeing and perception. 
Recent studies~\citep{zhang2025mllms, blink, liu2024seeing, xing2024causal, wang2025mllmseedynamiccorrection, kang2025see}
reveal that, while VLMs can often ``see'' relevant image regions for the question, they do not necessarily output correct answers. 
Training-free methods such as visual cropping~\cite{zhang2025mllms} have been proven to be an effective way to enhance fine-grained perception, but at the cost of significantly increased inference time. Additionally, visual cropping shows little improvement on questions involving relational reasoning and counting. 
Another alternative is to strengthen interaction between image and text by retraining the model~\cite{xing2024causal}, which requires a large amount of computational resources.

\begin{figure*}[tbp]
  \centering    
    \includegraphics[width=1.0\linewidth]{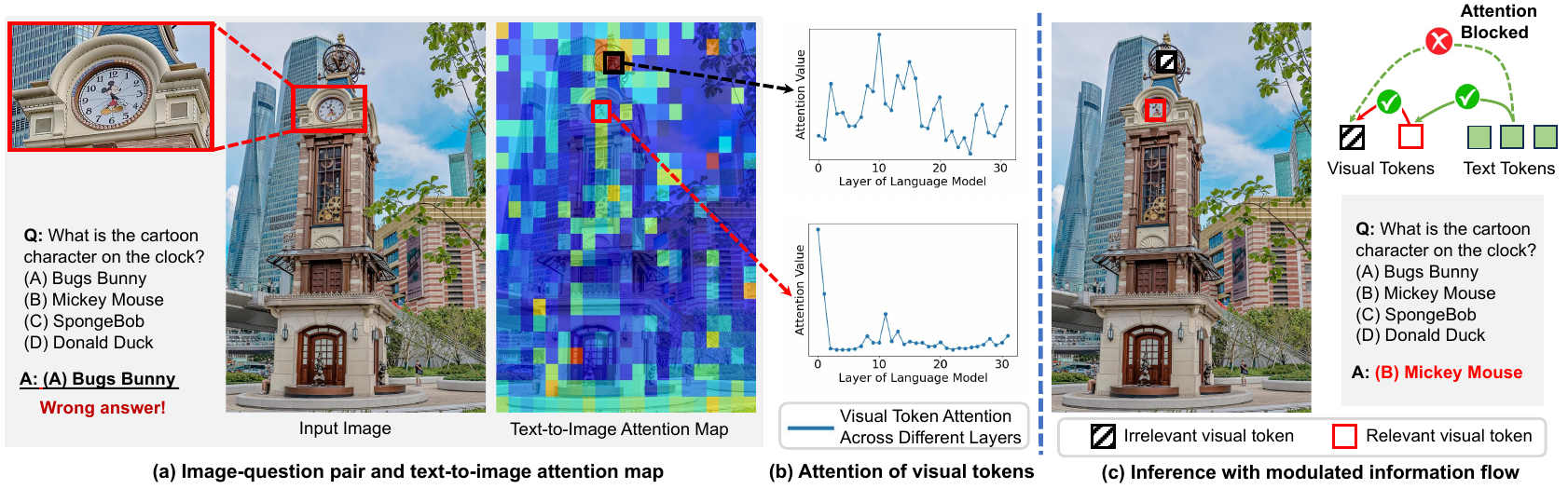}
    \caption{
    \textbf{Test-time information flow modulation improves perception.} 
    \textbf{(a)} VLMs often manage to capture related object regions with text-to-image attention (\eg, \emph{Mickey Mouse} clock in the input image), but sometimes fail to answer the question correctly. Here we use LLaVA-1.5~\cite{liu2024llava} as an example.
    \textbf{(b)} We observe that visual tokens corresponding to object regions show distinctive activation patterns in certain layers of the language model. In contrast, visual tokens in distracting regions exhibit irregular activation patterns across different layers.
    \textbf{(c)} Based on (b), we observe that disconnecting the interactions between text tokens and distracting visual tokens during inference leads to improved results, enabling the model to correctly select \emph{Mickey Mouse} for question answering. This information flow modulation is implemented by modulating the causal mask during attention computation.}    
  \label{fig:figure1}
\vspace{-1em}
\end{figure*}

In this work, we demonstrate that the misalignment between seeing and perception could be linked to suboptimal information flow~\cite{kaduri2025info,zhang2025info} during decoding. 
As shown in Fig.~\ref{fig:figure1}(a), instruction tokens spread their cross-attention across many background visual tokens, forming a spatially dispersed attention pattern. Instead of concentrating on the object regions that provide the necessary evidence for the answer, the attention mass is distributed over irrelevant visual tokens. This dispersion introduces noisy visual information into the textual decoding path, which has been shown to distract models and potentially lead to incorrect answers~\citep{chen2024fastv, zhang2025mllms, zhang2025fromred}.

Interestingly, we found that this misalignment could be addressed by modulating information flow during inference. 
To be specific, by blocking the attention between irrelevant visual tokens and text tokens, VLMs can re-route an effective pathway towards important image regions, producing correct answers. Fig.~\ref{fig:figure1}(c) illustrates an example of information flow modulation. 
During inference, the attention between irrelevant visual tokens and text tokens will be blocked. This enables the model to focus on important related regions and output the correct answer.

Based on the above observation, we propose to enhance the perception capability of VLMs via test-time information flow modulation, without the need for training.
The idea is that VLMs only need to focus on important tokens when answering the questions. As such, the key problem is how to identify important visual tokens. We observe that the attention statistics of visual tokens during decoding give hints about the importance of tokens. Fig.~\ref{fig:figure1}(b) shows that visual tokens corresponding to object regions are highly activated in certain layers, while irrelevant regions exhibit irregular activation patterns.
We denote the statistics of visual tokens as \emph{token dynamics}, as they record the change of tokens throughout the decoding process. Based on this, we develop a token dynamics-based entropy measurement to determine which visual tokens shall be retained and masked. 
Subsequently, visual token masking is efficiently implemented by causal mask modulation. With one additional decoding step to obtain and apply the modulated causal mask, the rest of the inference process remains the same. 

We comprehensively evaluate the proposed method on various tasks, including visual question answering (VQA)~\cite{vstar,blink,realworldqa}, visual grounding and counting~\cite{refcoco2014,yu2016refcoco,countbench}, optical character recognition (OCR)~\cite{textvqa,chartqa2022}, and object hallucination~\cite{li2023pope}.  
Experiments and analysis reveal some interesting findings: i) by knocking out the information flow between specified visual tokens and text tokens, we observe that only a subset of visual tokens has a significant impact on model output;
ii) information flow modulation is a powerful but underexplored tool to improve VLMs without additional training.

The main contributions are summarized as follows:

\begin{itemize}
    \item We show that modulating information flow during inference can improve the perception capability of VLMs.
    \item We propose a training-free, adaptive information flow (AIF) modulation method based on token dynamics.    
    \item Extensive experiments show that our approach notably improves the performance of two representative VLMs.
\end{itemize}

\section{Related Work}

\label{sec:formatting}

\vspace{-1em}
\paragraph{Vision-Language Models.}
VLMs have rapidly advanced across a wide range of applications, such as general perception~\citep{vstar, realworldqa, blink}, OCR~\citep{textvqa, chartqa2022}, visual grounding~\citep{refcoco2014}, multimodal embedding~\cite{kim2026wacv},
and automatic model discovery~\cite{jungmok2025automated}. However, hallucinations from text–vision mismatch remain common~\citep{li2023pope,yebin2024eccv,sungbin2025avhbench}. Data-centric instruction tuning can mitigate such errors but requires substantial retraining~\citep{liu2024llava, dai2023instructblip}, while verifier- or re-ranking-based pipelines improve factuality via external detectors or scorers at the cost of added latency and system complexity~\citep{zhao2025icml}. 
Complementary efforts target fine-detail perception: test-time cropping to magnify question-relevant regions~\citep{zhang2025mllms} and high-resolution or multi-tile inputs~\citep{bai2025qwen25, chen2024internvl, internvl3} improve small-object perception but increase compute or integration burden. In contrast, we leave weights and inputs unchanged and modulate the causal mask at inference time to disable connections between text tokens and irrelevant visual tokens, improving both fine-detail perception and factual alignment without retraining.

\paragraph{Information Flow.}
Information-flow analysis aims to localize where and how visual and linguistic evidence fuse and propagate across depth and tokens, often revealing hubs, relays, and token- or layer-level contributions. A consistent picture has emerged about where multimodal fusion happens and how it evolves with depth. Most prior work uses these analyses for three purposes: (i) diagnosis and interpretability, mapping layers and tokens that deliver vision to language and identifying flow cliffs where additional image tokens bring little benefit~\citep{kaduri2025info, zhang2025info, zhang2025fromred}; (ii) efficiency, pruning or compressing tokens to accelerate inference at the risk of losing informative cues~\citep{chen2024fastv, yin2025lifting}; and (iii) post-hoc output correction, reweighting or re-ranking scores after the forward pass instead of changing the internal pathways that produced them~\citep{li2023contrastivedecodingopenendedtext, vcd2024cvpr}. These approaches are largely descriptive or act after the fact and rarely alter intra-step connectivity within the same pass.
 Instead, we treat information flow as a control signal. 
 We use an entropy-based measurement to flag low-value visual tokens for the current query and analyze token dynamics across layers to identify where visual evidence stabilizes or pivots, as indicated by critical and mutation layers~\citep{wang2025knowledge}. 
 We then gate edges between text tokens and those low-value visual tokens in the causal mask at inference time. This pathway-level control preserves beneficial vision-to-vision aggregation while steering attention toward prompt-relevant visual evidence and strengthening general perception.

\paragraph{Causal Mask.}
Recent studies argue that the left-to-right causal masking inherited from language models can constrain cross-modal interaction in multimodal settings, motivating a redesign of the mask. \citet{pei2025rethinking} propose future-aware attention that lets vision tokens benefit from later textual cues while preserving autoregressive decoding. Similarly, \citet{wang2025seeing} establish a modality-mutual path during supervised fine-tuning, allowing vision tokens to attend to the question text. From a positional perspective, \citet{yin2024causal} relax the causal mask with pseudo-attention values to better encode positional information, while \citet{xing2024causal} redesign token layout with a concentric scheme to reduce the effective distance between visual and text tokens. These methods alter attention patterns via architectural, positional, or training-time changes that are baked into the model through pretraining or additional tuning.
In contrast, we keep the model architecture, weights, positional encodings, and sequence order unchanged and only modulate the causal mask at inference time. Guided by token-level dynamics, we selectively block connections from text tokens to irrelevant
visual tokens. This objective-aware, architecture-agnostic control requires no auxiliary modules or retraining while preserving helpful vision-to-vision aggregation and improving perception.

\section{A Closer Look at Text-Image Interaction in VLMs} \label{sec:motivation}

In this section, we investigate the interaction between visual tokens and text tokens, and study how visual-to-textual information flow affects model output. We start by reviewing the workflow of VLMs in Sec.~\ref{sec:preliminary}. Then, we quantify and analyze the dynamics of visual tokens during decoding process in Sec.~\ref{sec:dynamics}, showing that only a subset set of visual tokens contribute significantly to the output of large language model (LLM). Finally, we demonstrate that information flow modulation can improve the perception capability of VLMs in Sec.~\ref{sec:info_flow_matter}. These findings provide new insights into the information flow within VLMs.

\begin{figure}[tp]
  \centering
    \includegraphics[width=\linewidth]{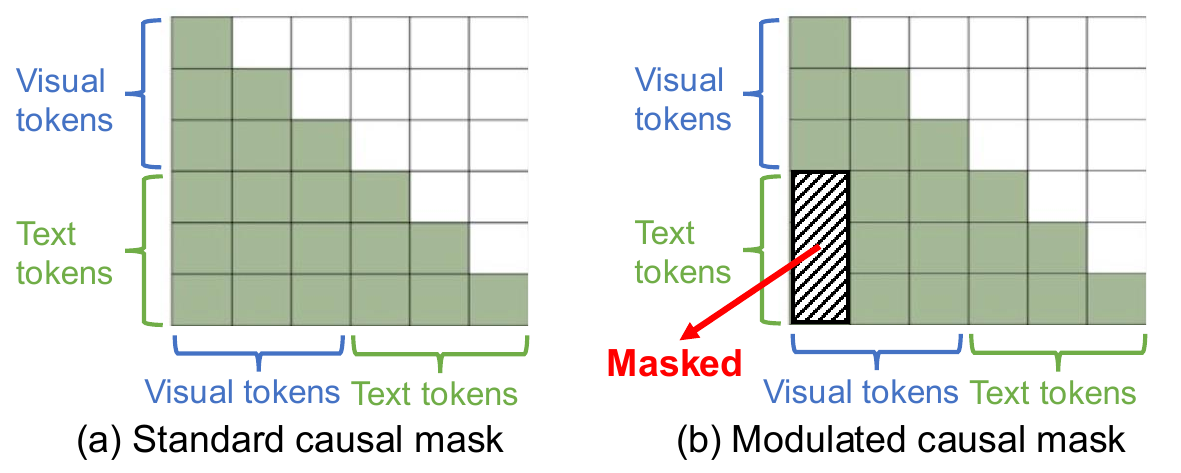}
    \caption{\textbf{Illustration of information flow modulation.} We only block the information flow between visual and text tokens. The information flow within visual tokens remains unchanged, ensuring that no image content is missing.}
  \label{fig:info_flow_mod}
\vspace{-1em}
\end{figure}

\subsection{Preliminary} \label{sec:preliminary}
Given the input image and text prompt, VLMs convert them into a set of visual tokens $\{ v_1, \ldots, v_M \}$. These tokens are concatenated with text tokens $\{t_1, \ldots, t_N \}$ and pass through the language model, outputting answers in an autoregressive manner. During language decoding, visual tokens and text tokens will interact with each other through attention mechanism~\cite{transformer2017} under the constraint of causal mask. 
Fig.~\ref{fig:info_flow_mod}(a) shows an example of causal mask $C \in \mathbb{R}^{(M+N)\times(M+N)}$.
For the $i$-th row in the causal mask, the values are initialized as:
\begin{equation}
C_{i,j} = 
\begin{cases}
    0, & \text{if } j \le i \\
    -\infty, & \text{otherwise}
\end{cases}.
\end{equation}
This causal masking strictly restricts the information flow during decoding: visual tokens can only attend to earlier visual tokens or prefixes of the text, and cannot benefit from textual cues that appear later in the instruction. Consequently, modulating the causal mask can directly reshape the cross-modal information flow and substantially influence the model’s overall performance.

\subsection{Dynamics of Visual Tokens} \label{sec:dynamics}
We investigate the interactions between visual and text tokens across different layers of LLM, which are referred to as token dynamics. 
We quantify token dynamics using image-to-text attention. Let $a_{i,j}^l$ denote the attention value between $i$-th visual token and $j$-th text token in LLM layer $l$. We define the token dynamics of visual token $v_i$ as follows:
\begin{equation}
    \mathcal{D}_{v_i}= \{d_{v_i}^l\}_{l=1}^{L}, \quad d_{v_i}^l = \max\nolimits_j a_{i,j}^l,
\end{equation}
where $L$ is the number of LLM layers and $j \in \{1, \ldots, N \}$. Token dynamics indicate how strongly each visual token attends to text tokens at different decoding stages. 
For convenience, we denote the average value in $\mathcal{D}_{v_i}$ as $\mu_{v_i}$:
\begin{equation}
    \mu_{v_i} = \frac{1}{L} \sum\nolimits_{l=1}^L d_{v_i}^l.
\end{equation}

\paragraph{Experiment Setup.}
To explore the relationship between token dynamics and LLM output, we perform analysis with LLaVA-1.5-7B~\cite{liu2024llava} on multiple datasets, including visual question answering~\cite{realworldqa}, OCR understanding~\cite{textvqa}, and object counting~\cite{countbench}. We first perform one-step LLM decoding to collect $\mathcal{D}_{v_i}$ for each visual token. Then, we block the information flow between visual tokens and text tokens.
Two settings are adopted to evaluate the impact of visual tokens, including masking low attention and high attention tokens.
For instance, when masking low attention tokens with a mask ratio $r$, we first sort visual tokens based on $\mu_{v_i}$ and then mask $rM$ visual tokens with the smallest $\mu_{v_i}$ values.
We vary the masking ratio and evaluate the performance.

Fig.~\ref{fig:info_flow_mod}(b) shows an example of the modulated causal mask, where one visual token is masked. Note that the remaining visual tokens still have access to the masked one, ensuring that image information is preserved. 
This stands in clear contrast to visual token pruning~\cite{chen2024fastv}, where the pruned tokens are removed from the sequence entirely, preventing both visual and text tokens from attending to them and thereby discarding their visual information.

\begin{figure}[tp]
  \centering
    \includegraphics[width=\linewidth]{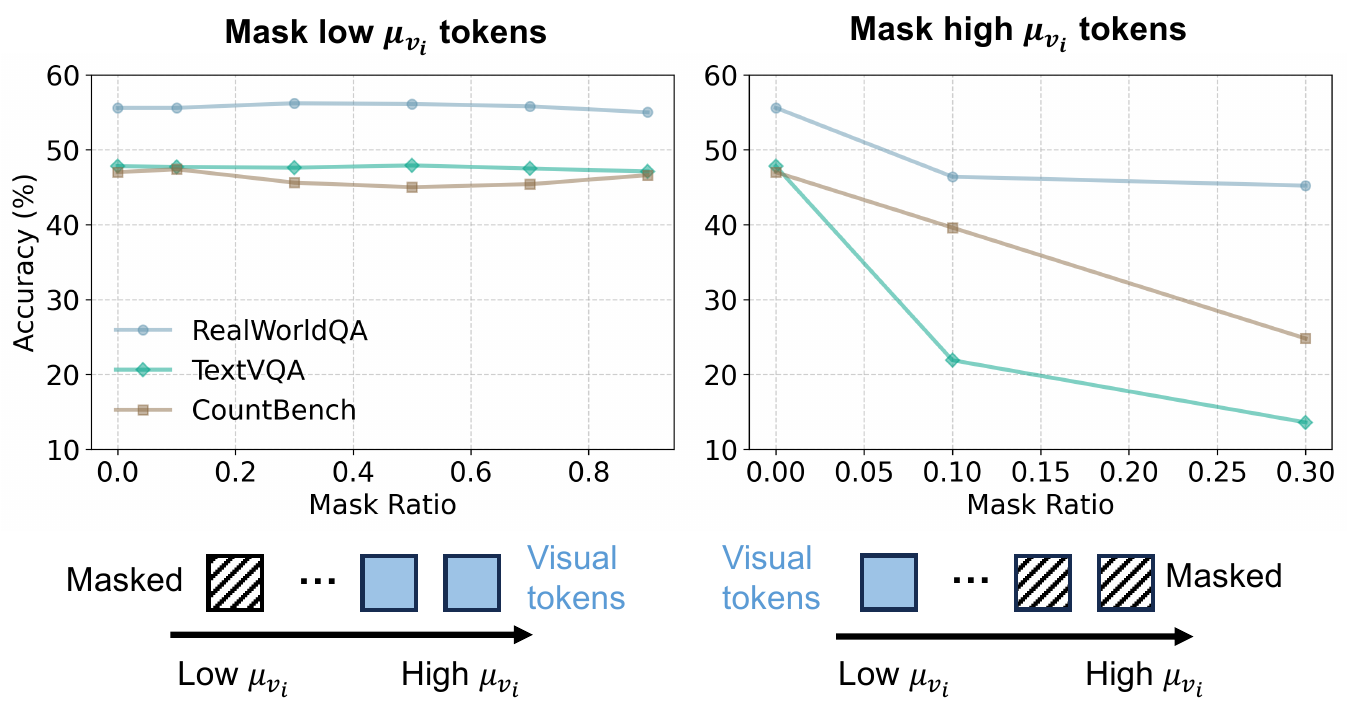}
    \caption{\textbf{Results of information flow modulation.} Visual tokens with low $\mu_{v_i}$ have a little impact on the performance, while masking high $\mu_{v_i}$ tokens considerably degrades the performance. This indicates the importance of high $\mu_{v_i}$ tokens.}    
  \label{fig:info_flow_exp}
\vspace{-0.5em}
\end{figure}

\begin{figure}[tp]
  \centering
\includegraphics[width=\linewidth]{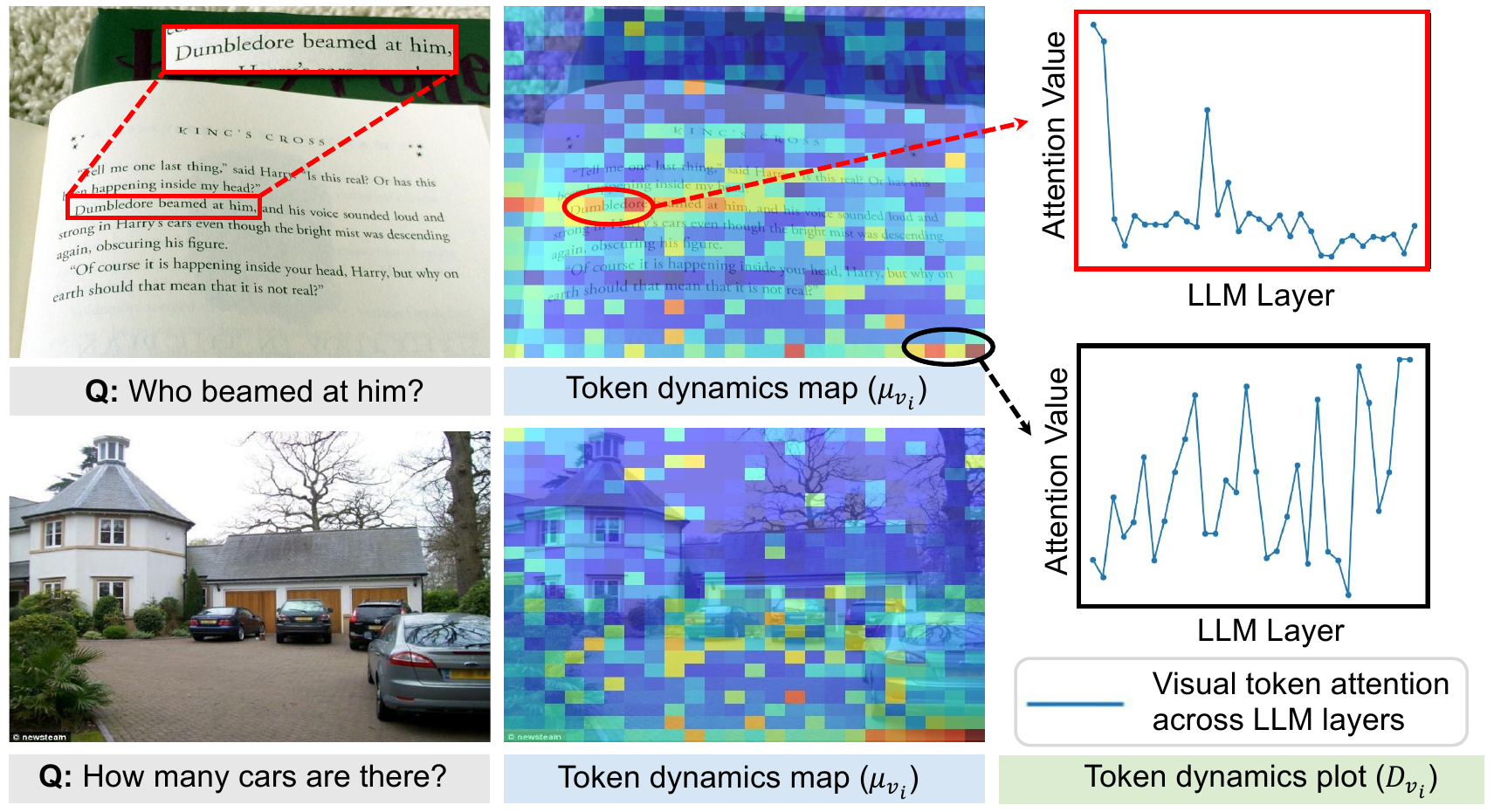}
    \caption{\textbf{Visualizations of token dynamics.}}
  \label{fig:token_dynamics}
\vspace{-1.5em}
\end{figure}

\paragraph{Results.} Figure~\ref{fig:info_flow_exp} shows the results of information flow modulation on different datasets. One can observe that: i) visual tokens with low $\mu_{v_i}$ values have little impact on the final performance, and the performance remains stable even with $90\%$ mask ratio; ii) visual tokens with high $\mu_{v_i}$ values significantly affect model performance, \eg, blocking the information flow with $10\%$ mask ratio leads to over $50\%$ relative performance drop on the TextVQA dataset; iii) different datasets show varied sensitivity to mask ratios. As shown in Fig.~\ref{fig:token_dynamics}, OCR understanding is a fine-grained task, often requiring precise perception of small objects. Masking visual tokens with high $\mu_{v_i}$ values will lead to information loss. In contrast, images with questions related to medium and large objects are less affected by masking.

\paragraph{Insights.} The above results reveal that only a subset of visual tokens have a significant impact on model outputs.
Since token dynamics capture how strongly each visual token attends to instruction tokens across layers, they serve as an effective indicator of the token's downstream influence. Consequently, the measured token dynamics provide a reasonable proxy for estimating each token's overall impact.

\subsection{Visual-to-Textual Information Flow Matters} \label{sec:info_flow_matter}

Section~\ref{sec:dynamics} shows that only a subset of visual tokens has a significant impact on the output of LLM. 
Interestingly, we observe that selectively masking some of these visual tokens sometimes improves the results. As such, we further investigate how much gain can be obtained by manually modulating the information flow between visual tokens and text tokens.
It is important to note that this experiment is not part of our proposed method. Rather, it serves as an oracle analysis designed to measure the potential gains of manual information flow modulation.

\paragraph{Experiment Setup.} We conduct experiments on RealWorldQA~\cite{realworldqa}, TextVQA~\cite{textvqa}, and CountBench~\cite{countbench} datasets.
We evaluate LLaVA-1.5-7B~\cite{liu2024llava} under standard causal mask and modulated causal mask settings. 
For the modulated causal mask setting, we sort visual tokens according to $\mu_{v_i}$ and use different mask ratios (\eg, from $0.1$ to $0.9$ with an interval of $0.1$) to mask visual tokens, resulting in a set of causal masks.
For each image, if there exist causal mask settings that enable the model to output the same answer as ground-truth, we record this image as correct when computing the accuracy. This quantifies the potential gains achieved by modulating information flow.

\begin{table}[t]
    \centering
	\caption{Results of manual information flow modulation on LLaVA-1.5-7B~\cite{liu2024llava}.}\vspace{-1mm}
		\setlength{\tabcolsep}{4pt}
        \resizebox{1.0\linewidth}{!}{
		\begin{tabular}{@{}l| c c c}
			\toprule               
                Causal Mask & RealWorldQA & TextVQA & CountBench \\
                \midrule                
                Standard & 55.6 & 47.8 & 47.0 \\                 
                Modulated & 61.6 & 49.9 & 50.3 \\
            \bottomrule
	    \end{tabular}}
	\label{table:manual_info_flow}
    \vspace{-1em}
\end{table}

\begin{figure}[tbp]
  \centering
    \includegraphics[width=\linewidth]{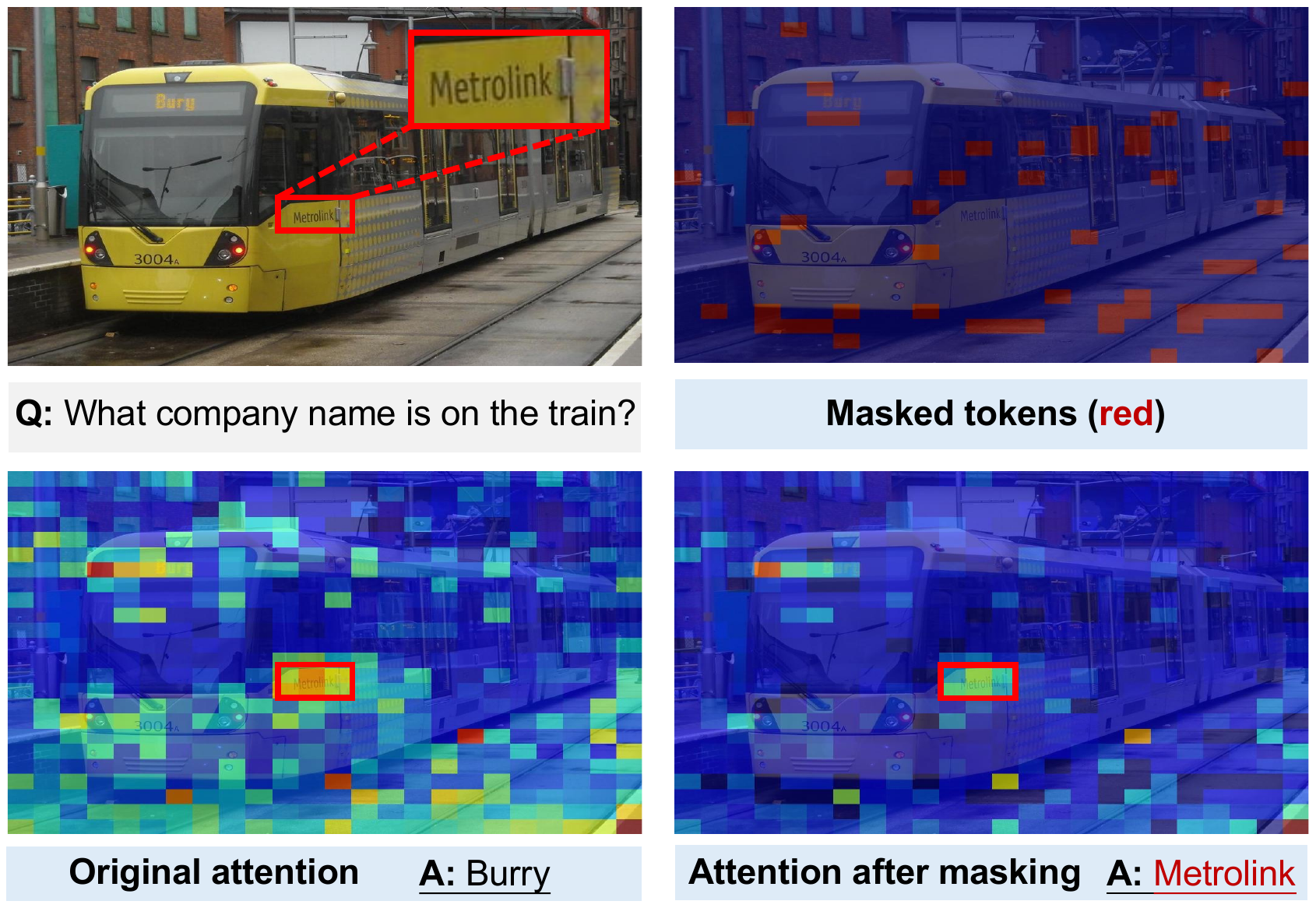}
    \caption{\textbf{Visualizations of attention and prediction with standard causal mask and modulated causal mask.}}
  \label{fig:modulated_causal_mask}
\vspace{-1.5em}
\end{figure}

\begin{figure*}[!htbp]
  \centering
    \includegraphics[width=0.9\linewidth]{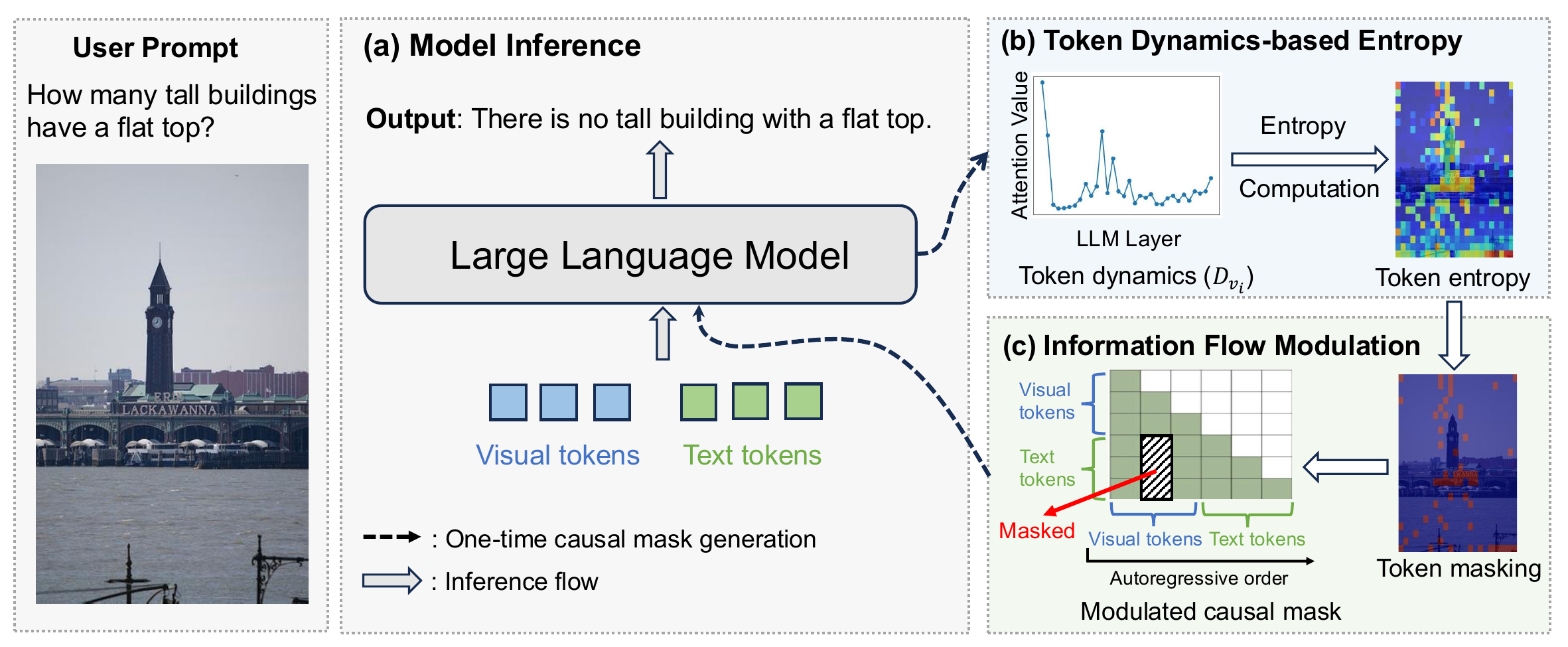}\vspace{-2mm}    
    \caption{\textbf{Pipeline overview}. Given an input image and user prompt, one-step language decoding is applied to obtain \textbf{(b)} token dynamics. Then, token dynamics-based entropy is computed, resulting in a token entropy map. Next, \textbf{(c)} adaptive token masking is adopted for causal mask generation. The generated causal mask is fed to the language model for reasoning, achieving test-time information flow modulation.}
  \label{fig:pipeline}
\vspace{-1.5em}
\end{figure*}

\paragraph{Results.} Table~\ref{table:manual_info_flow} reports the results of standard causal mask and modulated causal mask. It is clear that modulating visual-to-textual information flow can improve the results. The reason is that irrelevant regions are masked, enabling VLMs to focus on object regions related to the question. An example is shown in Fig. \ref{fig:modulated_causal_mask}. With a standard causal mask, the model incorrectly answers the question with ``Burry''. After applying the modulated causal mask, the attention gathers around related image regions, enabling the model to answer with ``Metrolink''.
\vspace{-0.5em}

\paragraph{Insights.} These results demonstrate that the quality of visual-to-textual information flow, not merely the presence of visual tokens, critically shapes the model's final prediction. By suppressing distracting or irrelevant visual tokens, the model is forced to align its textual reasoning with genuinely informative regions, leading to more grounded and consistent answers. This suggests that information flow structure itself is an underexplored but powerful lever for improving VLM perception, pointing toward new approaches that go beyond adding more data or architectural complexity and instead directly regulate how modalities interact inside the model.

\section{Perception with Adaptive Information Flow} 

Motivated by the success of information flow modulation (Sec.~\ref{sec:motivation}), we propose an adaptive information flow method for improving the perception capability of VLMs. In this section, we first give an overview of the proposed method. Then, we demonstrate how to measure the importance visual tokens via token dynamics. Lastly, we show that the information flow is modulated through causal mask.

\subsection{Overview}

The pipeline of our approach is depicted in Fig.~\ref{fig:pipeline}. The input image and user prompt will be encoded as visual tokens and text tokens, respectively. These tokens are passed through a language model, producing internal attention statistics. Based on the attention values, we compute token dynamics-based entropy for each visual token, resulting in a token entropy map. Subsequently, adaptive masking is applied to determine what visual tokens should be masked. The interaction between these masked tokens and text tokens are blocked in the causal mask. Finally, the modulated causal mask is fed to the language model, outputting final answers. Note that the causal mask generation only requires one additional decoding process.

\subsection{Measuring the Importance of Visual Tokens}

As mentioned earlier, visual tokens corresponding to object regions and irrelevant regions exhibit distinct token dynamics. The former typically yields high attention value at certain LLM layers, while the latter is irregularly activated across all layers. In another words, randomness is the main difference between important and irrelevant visual tokens. As such, we propose to use entropy as a measurement to quantify the importance of visual tokens. Based on token dynamics, we compute the entropy of each visual token $v_i$ as follows: 
\begin{equation}    
    \texttt{Ent}_{v_i} = \sum_{l=1}^{L} - \frac{ d_{v_i}^l}{L\cdot \mu_{v_i}} \log(\frac{ d_{v_i}^l}{L\cdot \mu_{v_i}}).
    \label{eq:token_entropy}
\end{equation}
A high $\texttt{Ent}_{v_i}$ value implies high randomness, indicating that this visual token is of less importance. Therefore, high entropy tokens will be masked first.

\subsection{Adaptive Information Flow Modulation}

After computing the entropy of visual tokens, the next step is to decide which tokens should be masked to modulate information flow. 
Inspired by the results in Sec.~\ref{sec:info_flow_matter} and the observation in Fig.~\ref{fig:modulated_causal_mask}, we expect the spatial distribution of visual-to-textual attention to be centralized to certain object regions after masking. In this case, the randomness of attention distribution should decrease. Therefore, we continue to use entropy as a measurement to quantify the shift in attention distribution.

Technically, given a set of visual tokens, we quantify the attention distribution as follows:
\begin{equation}\label{eq:att_entropy}
    S_0 = \sum_{i=1}^{M} - \frac{ \mu_{v_i}}{\sum_i  \mu_{v_i}} \log(\frac{ \mu_{v_i}}{\sum_i  \mu_{v_i}}).
\end{equation}
After applying token masking, we anticipate that the attention distribution of retained visual tokens differs from $S_0$. To simplify the masking process, we formulate it as an optimal thresholding problem. Given a set of candidate mask ratios (\eg, from $0.1$ to $0.9$ with an interval of $0.1$), we mask visual tokens in order based on the value $\texttt{Ent}_{v_i}$. Then, we quantify the attention distribution following Eq.~\ref{eq:att_entropy} for each mask ratio. 
The mask ratio with entropy value that is most distant from $S_0$ is selected for token masking.
Subsequently, the interaction between these masked visual tokens and text tokens will be blocked in causal mask. The information flow modulation is implemented by passing through the modulated causal mask to LLM (Fig.~\ref{fig:pipeline}(c)). 

\section{Results and Discussion}

\subsection{Datasets and Implementation Details}

\paragraph{Implementation Details.} We apply our approach to two popular VLMs, including LLaVA-1.5~\cite{liu2024llava} and Qwen2.5-VL~\cite{bai2025qwen25}. For LLaVA-1.5, we set the prompt following ViCrop~\cite{zhang2025mllms}. For Qwen2.5-VL, unless otherwise specified, we adopt the default prompt provided by VLMEvalKit~\cite{vlmevalkit}. Details can be found in the supplementary material.

\paragraph{Datasets.} We evaluate our approach across a wide range of tasks, including general visual question answering~\cite{vstar,realworldqa,mmstar2024}, OCR understanding~\cite{textvqa,seedbench}, visual grounding and counting~\cite{refcoco2014,yu2016refcoco,nagaraja2016refcoco}, and hallucination~\cite{li2023pope}.

\paragraph{Evaluation Metrics.} We follow VLMEvalKit~\cite{vlmevalkit} to compute the official metric for each dataset. For the hallucination dataset, we follow LLaVA-1.5~\cite{liu2024llava} to evaluate the method on three sampled subsets of COCO~\cite{coco2014}, and we report averaged accuracy across three splits.

\begin{table*}[t]
    \centering
    \caption{Comparisons with state-of-the-art methods on various benchmarks. The results of GPT-4o and Claude-3.5 Sonnet are from~\cite{bai2025qwen25}.}
    \setlength{\tabcolsep}{6pt}
		\begin{tabular}{@{}l l l l l l l l l @{}}
		\toprule
        \multirow{2}{*}{Method} & \multicolumn{3}{c}{\textbf{General VQA}} & \multicolumn{2}{c}{\textbf{OCR Understanding}} & \textbf{Counting}\\ 
        \cmidrule(lr){2-4} \cmidrule(lr){5-6} \cmidrule(lr){7-7} \cmidrule(lr){8-8}      
        & V* & RealWorldQA & MMStar & $\text{TextVQA}_{\text{val}}$ & SeedBench2-Plus & CountBench \\
        \midrule
        GPT-4o~\cite{gpt4o} & 73.9 & 75.4 & 64.7 & 77.4 & 72.0 & 87.9 \\
        Claude-3.5 Sonnet~\cite{claude3_5} & - & 60.1 & 65.1 & 76.5 & 71.7 & 89.7  \\        
        \midrule
        Qwen2.5-VL-72B~\cite{bai2025qwen25} & 86.4 & 75.7 & 70.8 & 83.5 & 73.0 & 93.6 \\
        \midrule
        LLaVA-1.5-7B~\cite{liu2024llava}  & 42.4 & 55.6 & 33.1 & 47.8 & 41.3 & 47.0 \\        
        \rowlightgray 
        LLaVA-1.5-7B AIF (\textbf{Ours}) & 50.3 \greensub{+7.9} & 60.5 \greensub{+4.9} & 39.5 \greensub{+6.4} & 49.9 \greensub{+2.1} & 44.9 \greensub{+3.6} & 50.1 \greensub{+3.1}  \\
        Qwen2.5-VL-7B~\cite{bai2025qwen25}  & 78.5 & 68.5 & 63.9 & 84.9 & 
        70.4 & 87.1 \\
        \rowlightgray 
        Qwen2.5-VL-7B AIF (\textbf{Ours}) & 84.8 \greensub{+6.3} & 74.5 \greensub{+6.0} & 70.9 \greensub{+7.0} & 86.0 \greensub{+1.1} & 76.5 \greensub{+6.1} & 89.5 \greensub{+2.4} \\        
        \bottomrule
	    \end{tabular}
	\label{table:vqa}
    \vspace{-0.8em}
\end{table*}

\subsection{Main Results}

\paragraph{General Visual Question Answering.} We first evaluate our method on general visual answering datasets, including V*~\cite{vstar}, RealWorldQA~\cite{realworldqa}, and MMStar~\cite{mmstar2024}. These datasets cover a wide range of real-world scenarios and applications, such as visual perception, logical reasoning, science question answering, and mathematics. Table~\ref{table:vqa} reports the results. Our approach substantially improves the performance of LLaVA-1.5 and Qwen2.5-VL. This indicates that our approach can improve the results by leveraging the information in the question.

\paragraph{OCR Understanding.} OCR-related tasks often involve fine-grained recognition, where the model needs to produce detailed answers instead of a single option letter. We evaluate our method on two OCR understanding datasets, including TextVQA~\cite{textvqa} and SeedBench2-Plus~\cite{seedbench} datasets. Table~\ref{table:vqa} shows that our approach consistently improves the performance of baselines. This supports the effectiveness of our approach in fine-grained tasks. 

\begin{table*}[!htbp]
    \centering
	\caption{Comparisons on visual grounding datasets.}
		\setlength{\tabcolsep}{3.2pt}
        \begin{tabular}{l| l l l| l l l| l l| l}
			\toprule
                 \multirow{2}{*}{Method} & \multicolumn{3}{c|}{\textbf{RefCOCO}} 
                 & \multicolumn{3}{c|}{\textbf{RefCOCO+}}                
                 & \multicolumn{2}{c|}{\textbf{RefCOCOg}} & \multirow{2}{*}{\textbf{Overall}} \\                
                & Val & TestA & TestB & Val & TestA & TestB & Val & Test \\
                \midrule
                Grounding-DINO-L~\cite{gdino2024eccv} & 90.6 & 93.2 & 88.2 & 82.8 & 89.0 & 75.9 & 86.1 & 87.0 & 86.6 \\
                Localization Heads~\cite{kang2025locheads} & 87.2 & 90.0 & 83.3 & 82.7 & 88.5 & 74.0 & 84.3 & 85.5 & 84.4 \\
                \midrule                
                \midrule
                Shikra-7B~\cite{chen2023shikra} & 87.0 & 90.6 & 80.2 & 81.6 & 87.4 & 72.1 & 82.3 & 82.2 & 82.9 \\
                InternVL2-8B~\cite{chen2024internvl2} & 87.1 & 91.1 & 80.7 & 79.8 & 87.9 & 71.4 & 82.7 & 82.7 & 82.9 \\ 
                InternVL-2.5-8B~\cite{intervl2_5} & 90.3 & 94.5 & 85.9 & 85.2 & 91.5 & 78.8 & 86.7 & 87.6 & 87.6 \\
                Qwen2.5-VL-7B~\cite{bai2025qwen25} & 90.0 & 92.5 & 85.4 & 84.2 & 89.1 & 76.9 & 87.2 & 87.2 & 86.6 \\
                \rowlightgray Qwen2.5-VL-7B AIF (\textbf{Ours}) & 91.6 \greensub{+1.6} & 94.2 \greensub{+1.7} & 88.4 \greensub{+3.0} & 86.6 \greensub{+2.4} & 91.0 \greensub{+1.9} & 80.5 \greensub{+3.6} & 89.3 \greensub{+2.1} & 89.7 \greensub{+2.5} & 88.9 \greensub{+2.3} \\           
            \bottomrule
	    \end{tabular}
	\label{table:refcoco}	
\vspace{-1.4em}
\end{table*}

\paragraph{Grounding and Counting.} Visual grounding demands precise object identification and localization given specified textual descriptions. We compare with recent methods on RefCOCO~\cite{refcoco2014,yu2016refcoco,nagaraja2016refcoco} datasets. As shown in Table~\ref{table:refcoco}, our approach considerably improves the grounding results of Qwen2.5-VL, with an average improvement of $2.3$. It is worth noticing that our approach also outperforms Grounding-DINO~\cite{gdino2024eccv}, a specialized visual grounding method. This validates that our approach can improve precise object localization. 
Additionally, we also report the results on a counting dataset~\cite{countbench}. Our approach achieves an improvement of approximately $2-3\%$ over the baselines. 

\paragraph{Hallucination.} One may wonder whether modulating causal mask may lead to hallucination. As such, we further evaluate the proposed method on an object hallucination dataset~\cite{li2023pope}. Table~\ref{table:pope} shows that our approach continues to advance the performance of baselines. The results imply that information flow modulation is helpful for alleviating hallucinations, as VLMs can focus on related regions.

\paragraph{Comparisons with Additional Methods.} We further compare with a training-free~\cite{zhang2025mllms} and a training-based method~\cite{xing2024causal}. ViCrop~\cite{zhang2025mllms} introduces question-aware visual cropping to improve performance, while CCA~\cite{xing2024causal} adopts concentric causal attention to enhance the interaction between image and text. Table~\ref{table:comparison} shows the comparisons. Our approach outperforms these competing methods on general visual question answering and hallucination datasets without re-training, showcasing the superiority of our method.

\begin{table}[tbp]
    \centering
	\caption{Results on the POPE dataset.}
		\setlength{\tabcolsep}{6pt}
		\begin{tabular}{@{}l l| c}
			\toprule
                 Model & Decoding & Accuracy \\                 
                \midrule
                \multirow{2}{*}{LLaVA-1.5-7B~\cite{liu2024llava}} & Baseline & 85.4 \\                
                & \textbf{Ours} & 88.7 \\                
                \midrule
                \multirow{2}{*}{Qwen2.5-VL-7B~\cite{bai2025qwen25}} & Baseline & 87.8 \\
                & \textbf{Ours} & 89.5 \\                            
            \bottomrule
	    \end{tabular}
	\label{table:pope}	
\vspace{-0.7em}
\end{table}

\begin{table}[tbp]
    \centering
	\caption{Comparisons with additional methods. All methods use LLaVA-1.5-7B as the base model.}
		\setlength{\tabcolsep}{3pt}
		\begin{tabular}{@{}l| c| c c c c}
			\toprule
                Method & Re-Training & MMStar & RealWorldQA & POPE \\ 
                \midrule
                LLaVA-1.5 & - & 33.1 & 55.6 & 85.4 \\                
                \midrule                
                CCA~\cite{xing2024causal} & \checkmark & 33.2 & 51.8 & 86.9 \\
                \midrule                
                ViCrop~\cite{zhang2025mllms} & \ding{56} & 35.2 & 56.7 & 87.2 \\       
                \textbf{Ours} & \ding{56} & \textbf{39.5} & \textbf{60.5} & \textbf{88.7} \\                
            \bottomrule
	    \end{tabular}
	\label{table:comparison}	
\vspace{-1.5em}
\end{table}

\begin{figure*}[!htbp]
  \centering
\includegraphics[width=\linewidth]{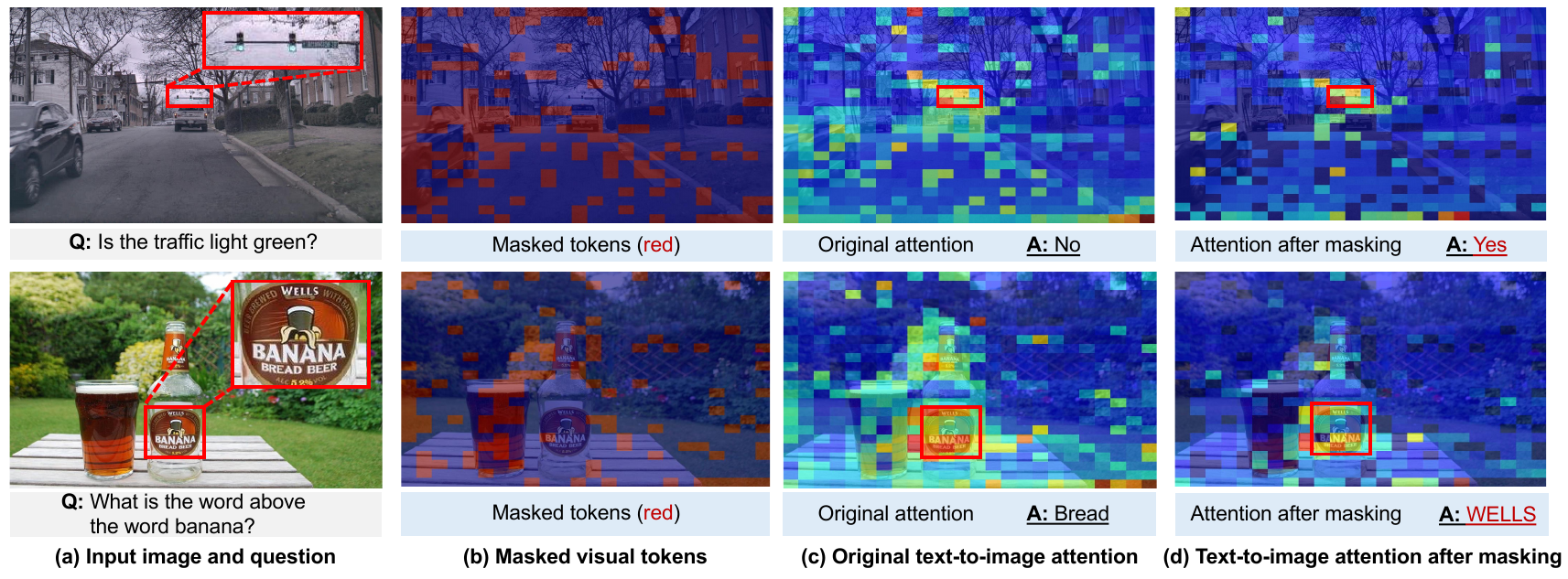}
    \caption{\textbf{Visualizations of masked tokens and text-to-image attention after information flow modulation.}}
  \label{fig:ablation_mask}
\vspace{-1.2em}
\end{figure*}

\begin{table}
    \centering
	\caption{Given selected top $20\%$ visual tokens, we compute the recall of object tokens across various object sizes on RefCOCO.}
		\setlength{\tabcolsep}{3pt} 
		\begin{tabular}{@{}l| l l l}
			\toprule
            \multirow{2}{*}{Token Selection Method} & \multicolumn{3}{c}{Object Size} \\ 
            & Small & Medium  & Large \\
            \midrule                                 
            High-Attention Tokens & 28.8 & 25.1 & 21.2 \\    
            Low-Entropy Tokens (\textbf{Ours}) & 49.7\greensub{+20.9} & 41.4\greensub{+16.3} & 31.0\greensub{+9.8} \\
            \bottomrule
	    \end{tabular}
	\label{table:attention_iou}	
    \vspace{-1.7em}
\end{table}

\subsection{Ablation Study}

Here we justify the design choices and effectiveness of our approach.
More results and discussions can be found in the supplementary material.

\paragraph{A Quantitative Justification of Token Entropy.}
To justify the effectiveness of token entropy (Eq.~\ref{eq:token_entropy}), we conduct a quantitative analysis on RefCOCO with Qwen2.5-VL-7B.
We first select the top $20\%$ tokens based on attention value and token entropy, respectively. Given these tokens, we compute the recall of object tokens and report results across different object sizes for clarity. Table~\ref{table:attention_iou} shows that:
i) $28.8\%$, $25.1\%$, and $21.2\%$ of object tokens from small, medium, and large objects fall within the top $20\%$ of high-attention tokens. This suggests that VLMs could reasonably ``see'' the objects, consistent with findings in~\cite{liu2024seeing,zhang2025mllms}. Note that recall reflects only object-awareness and is not directly comparable to object detectors.
ii) our entropy-based method significantly improves recall across all object sizes, indicating that token entropy is a more reliable proxy for token importance than attention value. Note that the decrease in recall for large objects is reasonable, as the proportion of object tokens may exceed $20\%$.

\paragraph{What Visual Tokens Are Masked?} Fig.~\ref{fig:ablation_mask}(b) shows the visual tokens masked by our approach. As we can see, these masked tokens correspond to background or irrelevant objects. By blocking the information flow between these non-important tokens and text tokens, the model can focus on related object regions and output correct answers (Fig.~\ref{fig:ablation_mask}(d)).

\paragraph{Attention Distribution After Modulation.} We also visualize text-to-image attention before and after applying information flow modulation. As shown in Fig.~\ref{fig:ablation_mask}(c), object and background regions both show high attention values under standard causal mask setting. After applying information flow modulation, the attention centralizes to object regions as shown in Fig.~\ref{fig:ablation_mask}(d), leading to improved answers. 

\begin{table}[tbp]
    \centering
    \caption{Ablation on masking strategy.}
		\setlength{\tabcolsep}{3.5pt}        
		\begin{tabular}{@{}l| c c c c c}
			\toprule                
                Masking Setting & V* & RealWorldQA & CountBench \\ 
                \midrule      
                Random Token & 40.3  & 55.7  & 45.6  \\
                Low-Entropy Token & 38.7 & 53.2 & 45.2 \\
                \midrule
                \textbf{Ours} &  50.3 & 60.5 &  50.1 \\
            \bottomrule
	    \end{tabular}
	\label{table:random_mask}	
    \vspace{-0.7em}
\end{table}

\paragraph{Ablation on Masking Strategy.} We further compare our method with random token masking and low-entropy token masking under the same masking ratio. Table~\ref{table:random_mask} shows that: i) masking low-entropy visual tokens (\ie, important tokens) significantly degrades the performance; ii) random masking underperforms our approach. These results justify the effectiveness of our masking strategy.

\begin{table}[tbp]
    \centering
	\caption{Ablation on different types of modulated causal mask.}
		\setlength{\tabcolsep}{5pt}
		\begin{tabular}{@{}l| c c c c}
			\toprule
                Dataset & Baseline & Vis2Vis & Vis2Text & Ours \\ 
                \midrule 
                RealWorldQA & 55.6 & 48.9 & 49.8 & \textbf{60.5} \\
                CountBench & 47.0 & 41.3 & 33.5 & \textbf{50.1} \\                
            \bottomrule
	    \end{tabular}
	\label{table:ablation_causal_mask}	
\vspace{-1.0em}
\end{table}

\paragraph{Comparison with Future-Aware Causal Mask.} Another way to modulate information flow is to adopt a future-aware causal mask~\cite{pei2025rethinking}.
We compare two variants: visual-to-visual (vis2vis) and visual-to-textual (vis2text). The former allows interaction between all visual tokens, and the latter enables each visual token to have access to text tokens.
Table~\ref{table:ablation_causal_mask} shows that these two types of causal masks deteriorate the performance. The reason is likely that future-aware causal mask significantly alters the attention distribution and makes negative impacts. In contrast, our approach consistently improves the performance.

\subsection{Discussion}

\paragraph{Computational Cost.} Our method consists of one LLM decoding step and causal mask generation. The computational cost of the LLM decoding step roughly equals to the time of one token generation. The causal mask generation step is computational efficient, which takes a few miliseconds. Overall, the inference cost of our method takes similar time as one token generation.

\paragraph{Limitation.} While our approach is model-agnostic and applicable to different kinds of tasks, the method may suffer from lengthy and indirect text prompt. In this case, it may be challenging for the model to capture important regions.

\section{Conclusion}
We have demonstrated that the perception capability of VLMs could be enhanced by test-time information flow modulation, where text tokens selectively attend to important visual tokens. Extensive experiments on diverse datasets validate the effectiveness and generality of our method. We hope our findings can spark new thinking about the interaction between image and text within VLMs.

\section*{Acknowledgment}

This work was supported by the KAIST Jang Young Sil Fellow Program, the InnoCORE program of the Ministry of Science and ICT (N10250156, KAIST InnoCore LLM), and the Institute of Information \& communications Technology Planning \& Evaluation (IITP) grant funded by the Korea government(MSIT) (No. RS-2024-00457882, National AI Research Lab Project; No. 2022-0-00124, No.RS-2022-II220124, Development of Artificial Intelligence Technology for Self-Improving Competency-Aware Learning; No.RS-2019-II191906, Artificial Intelligence Graduate School Program(POSTECH)).

{
    \small
    \bibliographystyle{ieeenat_fullname}
    \bibliography{main}

@String(CVPR  = {CVPR})

@String(ICCV  = {ICCV})

@String(ECCV  = {ECCV})

@String(NIPS  = {NeurIPS})

@String(ICML  = {ICML})

@String(ICLR  = {ICLR})

@String(EMNLP  = {EMNLP})

@String(NAACL  = {NAACL})

@String(WACV  = {WACV})

@String{arxiv 	= "arXiv"}

@InProceedings{kim2026wacv,
    author    = {Kim, Kyeong Seon and Seong-Eun, Baek and Jung-Mok, Lee and Oh, Tae-Hyun},
    title     = {m{EOL}: Training-Free Instruction-Guided Multimodal Embedder for Vector Graphics and Image Retrieval},
    booktitle = WACV,
    month     = {March},
    year      = {2026},
    pages     = {1191-1200}
}

@inproceedings{
jungmok2025automated,
title={Automated Model Discovery via Multi-modal \& Multi-step Pipeline},
author={Lee Jung-Mok and Nam Hyeon-Woo and Moon Ye-Bin and Junhyun Nam and Tae-Hyun Oh},
booktitle=NIPS,
year={2025},
url={https://openreview.net/forum?id=qGFvTIMS3W}
}

@InProceedings{yebin2024eccv,
author="Ye-Bin, Moon
and Hyeon-Woo, Nam
and Choi, Wonseok
and Oh, Tae-Hyun",
title="{BEAF}: Observing BEfore-AFter Changes to Evaluate Hallucination in Vision-Language Models",
booktitle=ECCV,
year="2024",
pages="232--248",
}

@inproceedings{
sungbin2025avhbench,
title={{AVHB}ench: A Cross-Modal Hallucination Benchmark for Audio-Visual Large Language Models},
author={Kim Sung-Bin and Oh Hyun-Bin and JungMok Lee and Arda Senocak and Joon Son Chung and Tae-Hyun Oh},
booktitle=ICLR,
year={2025},
url={https://openreview.net/forum?id=jTEKTdI3K9}
}

@inproceedings{
    kang2025see,
    title={See What You Are Told: Visual Attention Sink in Large Multimodal Models},
    author={Seil Kang and Jinyeong Kim and Junhyeok Kim and Seong Jae Hwang},
    booktitle=ICLR,
    year={2025},    
}

@INPROCEEDINGS{kang2025locheads,
  author={Kang, Seil and Kim, Jinyeong and Kim, Junhyeok and Hwang, Seong Jae},
  booktitle=CVPR, 
  title={Your Large Vision-Language Model Only Needs A Few Attention Heads For Visual Grounding}, 
  year={2025},  
  pages={9339-9350},
}

@InProceedings{vcd2024cvpr,
    author    = {Leng, Sicong and Zhang, Hang and Chen, Guanzheng and Li, Xin and Lu, Shijian and Miao, Chunyan and Bing, Lidong},
    title     = {Mitigating Object Hallucinations in Large Vision-Language Models through Visual Contrastive Decoding},
    booktitle = CVPR,
    month     = {June},
    year      = {2024},
    pages     = {13872-13882}
}

@inproceedings{zhao2025icml,
author = {Zhao, Linxi and Deng, Yihe and Zhang, Weitong and Gu, Quanquan},
title = {Mitigating object hallucination in large vision-language models via image-grounded guidance},
year = {2025},
booktitle = ICML,
pages = {77461 - 77486},
}

@InProceedings{tong2024mmvp,
    author    = {Tong, Shengbang and Liu, Zhuang and Zhai, Yuexiang and Ma, Yi and LeCun, Yann and Xie, Saining},
    title     = {Eyes Wide Shut? Exploring the Visual Shortcomings of Multimodal LLMs},
    booktitle = CVPR,
    month     = {June},
    year      = {2024},
    pages     = {9568-9578}
}

@inproceedings{mmbench,
author = {Liu, Yuan and Duan, Haodong and Zhang, Yuanhan and Li, Bo and Zhang, Songyang and Zhao, Wangbo and Yuan, Yike and Wang, Jiaqi and He, Conghui and Liu, Ziwei and Chen, Kai and Lin, Dahua},
title = {{MMBench}: Is Your Multi-modal Model an All-Around Player?},
year = {2024},
booktitle = ECCV,
pages = {216–233},
}

@InProceedings{kaduri2025info,
    author    = {Kaduri, Omri and Bagon, Shai and Dekel, Tali},
    title     = {What's in the Image? A Deep-Dive into the Vision of Vision Language Models},
    booktitle = CVPR,
    month     = {June},
    year      = {2025},
    pages     = {14549-14558}
}

@INPROCEEDINGS{zhang2025info,
  author={Zhang, Zhi and Yadav, Srishti and Han, Fengze and Shutova, Ekaterina},
  booktitle=CVPR, 
  title={Cross-modal Information Flow in Multimodal Large Language Models}, 
  year={2025},
  pages={19781-19791},
}

@inproceedings{yin2024causal,
author = {Yin, Qingyu and He, Xuzheng and Zhuang, Xiang and Zhao, Yu and Yao, Jianhua and Shen, Xiaoyu and Zhang, Qiang},
title = {Stable{M}ask: refining causal masking in decoder-only transformer},
year = {2024},
numpages = {20},
pages={57033--57052},
booktitle = ICML,
}

@inproceedings{xing2024causal,
  title={Mitigating object hallucination via concentric causal attention},
  author={Xing, Yun and Li, Yiheng and Laptev, Ivan and Lu, Shijian},
  booktitle=NIPS,
  pages={92012--92035},
  year={2024}
}

@article{wang2025seeing,
  title={Seeing is understanding: Unlocking causal attention into modality-mutual attention for multimodal llms},
  author={Wang, Wei-Yao and Wang, Zhao and Suzuki, Helen and Kobayashi, Yoshiyuki},
  journal={arXiv preprint arXiv: 2503.02597},
  year={2025}
}

@InProceedings{pei2025rethinking,
  title={Rethinking Causal Mask Attention for Vision-Language Inference},
  author={Pei, Xiaohuan and Huang, Tao and Ma, YanXiang and Xu, Chang},
  booktitle=ICLR,
  year={2026}
}

@article{intervl2_5,
  title={Expanding performance boundaries of open-source multimodal models with model, data, and test-time scaling},
  author={Chen, Zhe and Wang, Weiyun and Cao, Yue and Liu, Yangzhou and Gao, Zhangwei and Cui, Erfei and Zhu, Jinguo and Ye, Shenglong and Tian, Hao and Liu, Zhaoyang and others},
  journal={arXiv preprint arXiv: 2412.05271},
  year={2024}
}

@InProceedings{liu2024llava,
    author    = {Liu, Haotian and Li, Chunyuan and Li, Yuheng and Lee, Yong Jae},
    title     = {Improved Baselines with Visual Instruction Tuning},
    booktitle = CVPR,
    month     = {June},
    year      = {2024},
    pages     = {26296-26306}
}

@article{bai2025qwen25,
  title={Qwen2.5-{VL} technical report},
  author={Shuai Bai and Keqin Chen and Xuejing Liu and Jialin Wang and Wenbin Ge and Sibo Song and Kai Dang and Peng Wang and Shijie Wang and Jun Tang and Humen Zhong and Yuanzhi Zhu and Mingkun Yang and Zhaohai Li and Jianqiang Wan and Pengfei Wang and Wei Ding and Zheren Fu and Yiheng Xu and Jiabo Ye and Xi Zhang and Tianbao Xie and Zesen Cheng and Hang Zhang and Zhibo Yang and Haiyang Xu and Junyang Lin},
  journal={arXiv preprint arXiv: 2502.13923},
  year={2025},
}

@inproceedings{chen2024internvl,
  title={Intern{VL}: Scaling up vision foundation models and aligning for generic visual-linguistic tasks},
  author={Chen, Zhe and Wu, Jiannan and Wang, Wenhai and Su, Weijie and Chen, Guo and Xing, Sen and Zhong, Muyan and Zhang, Qinglong and Zhu, Xizhou and Lu, Lewei and others},
  booktitle=CVPR,
  pages={24185--24198},
  year={2024}
}

@article{guo2025seed,
  title={Seed1. 5-{VL} technical report},
  author={Guo, Dong and Wu, Faming and Zhu, Feida and Leng, Fuxing and Shi, Guang and Chen, Haobin and Fan, Haoqi and Wang, Jian and Jiang, Jianyu and Wang, Jiawei and others},
  journal={arXiv preprint arXiv: 2505.07062},
  year={2025}
}

@inproceedings{dai2023instructblip,
  title={Instruct{BLIP}: Towards general-purpose vision-language models with instruction tuning},
  author={Dai, Wenliang and Li, Junnan and Li, Dongxu and Tiong, Anthony and Zhao, Junqi and Wang, Weisheng and Li, Boyang and Fung, Pascale N and Hoi, Steven},
  booktitle=NIPS,
  pages={49250--49267},
  year={2023}
}

@inproceedings{transformer2017,
author = {Vaswani, Ashish and Shazeer, Noam and Parmar, Niki and Uszkoreit, Jakob and Jones, Llion and Gomez, Aidan N. and Kaiser, \L{}ukasz and Polosukhin, Illia},
title = {Attention is all you need},
year = {2017},
booktitle = NIPS,
pages = {6000–6010},
}

@article{seedbench,
  title={{SEED}-{B}ench-2-{P}lus: Benchmarking multimodal large language models with text-rich visual comprehension},
  author={Li, Bohao and Ge, Yuying and Chen, Yi and Ge, Yixiao and Zhang, Ruimao and Shan, Ying},
  journal={arXiv preprint arXiv: 2404.16790},
  year={2024}
}

@inproceedings{mmstar2024,
 author = {Chen, Lin and Li, Jinsong and Dong, Xiaoyi and Zhang, Pan and Zang, Yuhang and Chen, Zehui and Duan, Haodong and Wang, Jiaqi and Qiao, Yu and Lin, Dahua and Zhao, Feng},
 booktitle = NIPS,
 pages = {27056--27087},
 title = {Are We on the Right Way for Evaluating Large Vision-Language Models?},
 year = {2024}
}

@InProceedings{coco2014,
author="Lin, Tsung-Yi
and Maire, Michael
and Belongie, Serge
and Hays, James
and Perona, Pietro
and Ramanan, Deva
and Doll{\'a}r, Piotr
and Zitnick, C. Lawrence",
editor="Fleet, David
and Pajdla, Tomas
and Schiele, Bernt
and Tuytelaars, Tinne",
title="Microsoft {COCO}: Common Objects in Context",
booktitle=ECCV,
year="2014",
pages="740--755",
}

@INPROCEEDINGS{vstar,
  author={Wu, Penghao and Xie, Saining},
  booktitle=CVPR, 
  title={V*: Guided Visual Search as a Core Mechanism in Multimodal LLMs}, 
  year={2024},
  pages={13084-13094},
}

@inproceedings{blink,
author = {Fu, Xingyu and Hu, Yushi and Li, Bangzheng and Feng, Yu and Wang, Haoyu and Lin, Xudong and Roth, Dan and Smith, Noah A. and Ma, Wei-Chiu and Krishna, Ranjay},
title = {{BLINK}: Multimodal Large Language Models Can See but Not Perceive},
year = {2024},
booktitle = ECCV,
pages = {148–166},
}

@inproceedings{textvqa,
    title={Towards VQA Models That Can Read},
    author={Singh, Amanpreet and Natarjan, Vivek and Shah, Meet and Jiang, Yu and Chen, Xinlei and Batra, Dhruv and Parikh, Devi and Rohrbach, Marcus},
    booktitle=CVPR,
    pages={8317-8326},
    year={2019}
}

@inproceedings{vlmevalkit,
  title={{VLME}val{K}it: An open-source toolkit for evaluating large multi-modality models},
  author={Duan, Haodong and Yang, Junming and Qiao, Yuxuan and Fang, Xinyu and Chen, Lin and Liu, Yuan and Dong, Xiaoyi and Zang, Yuhang and Zhang, Pan and Wang, Jiaqi and Lin, Dahua and Chen, Kai},
  booktitle={ACM MM},
  pages={11198--11201},
  year={2024}
}

@misc{realworldqa,
  title = {Realworldqa: A benchmark for real-world spatial understanding},
    author={xAI},
  howpublished = {\url{https://huggingface.co/datasets/xai-org/RealworldQA}},
    year={2024},
}

@misc{gpt4o,
  title = {Hello gpt-4o},
    author={openAI},
  howpublished = {\url{https://openai.com/index/hello-gpt-4o}},
    year={2024},
}

@misc{claude3_5,
  title = {Claude 3.5 sonnet},
    author={Anthropic},
  howpublished = {\url{https://www.anthropic.com/news/claude-3-5-sonnet}},
    year={2024},
}

@InProceedings{countbench,
    author    = {Paiss, Roni and Ephrat, Ariel and Tov, Omer and Zada, Shiran and Mosseri, Inbar and Irani, Michal and Dekel, Tali},
    title     = {Teaching {CLIP} to Count to Ten},
    booktitle = ICCV,
    month     = {October},
    year      = {2023},
    pages     = {3170-3180}
}

@inproceedings{chartqa2022,
    title = "{C}hart{QA}: A Benchmark for Question Answering about Charts with Visual and Logical Reasoning",
    author = "Masry, Ahmed  and
      Long, Do  and
      Tan, Jia Qing  and
      Joty, Shafiq  and
      Hoque, Enamul",
    booktitle = "ACL Findings",    
    year = "2022",
    pages = "2263--2279",
}

@inproceedings{refcoco2014,
  title={{R}efer{I}t{G}ame: Referring to objects in photographs of natural scenes},
  author={Kazemzadeh, Sahar and Ordonez, Vicente and Matten, Mark and Berg, Tamara},
  booktitle={EMNLP},
  pages={787--798},
  year={2014}
}

@inproceedings{nagaraja2016refcoco,
  title={Modeling context between objects for referring expression understanding},
  author={Nagaraja, Varun K and Morariu, Vlad I and Davis, Larry S},
  booktitle=ECCV,
  pages={792--807},
  year={2016},
}

@inproceedings{yu2016refcoco,
  title={Modeling context in referring expressions},
  author={Yu, Licheng and Poirson, Patrick and Yang, Shan and Berg, Alexander C and Berg, Tamara L},
  booktitle=ECCV,
  pages={69--85},
  year={2016},
}

@inproceedings{li2023pope,
  title={Evaluating Object Hallucination in Large Vision-Language Models},
  author={Yifan Li and Yifan Du and Kun Zhou and Jinpeng Wang and Wayne Xin Zhao and Ji-rong Wen},
  booktitle=EMNLP,
  pages = "292--305",
  year={2023},
}

@inproceedings{
  zhang2025mllms,
  title={{MLLM}s Know Where to Look: Training-free Perception of Small Visual Details with Multimodal {LLM}s},
  author={Jiarui Zhang and Mahyar Khayatkhoei and Prateek Chhikara and Filip Ilievski},
  booktitle=ICLR,
  year={2025},
}

@inproceedings{gdino2024eccv,
author = {Liu, Shilong and Zeng, Zhaoyang and Ren, Tianhe and Li, Feng and Zhang, Hao and Yang, Jie and Jiang, Qing and Li, Chunyuan and Yang, Jianwei and Su, Hang and Zhu, Jun and Zhang, Lei},
title = {Grounding {DINO}: Marrying DINO with Grounded Pre-training for Open-Set Object Detection},
year = {2024},
booktitle = ECCV,
pages = {38–55},
}

@article{internvl3,
  title={Intern{VL}3: Exploring advanced training and test-time recipes for open-source multimodal models},
  author={Zhu, Jinguo and Wang, Weiyun and Chen, Zhe and Liu, Zhaoyang and Ye, Shenglong and Gu, Lixin and Tian, Hao and Duan, Yuchen and Su, Weijie and Shao, Jie and others},
  journal={arXiv preprint arXiv: 2504.10479},
  year={2025}
}

@inproceedings{zhang2025fromred,
  title={From Redundancy to Relevance: Information Flow in LVLMs Across Reasoning Tasks},
  author={Zhang, Xiaofeng and  Quan, Yihao and Shen, Chen and Yuan, Xiaosong and Yan, Shaotian and Xie, Liang and Wang, Wenxiao and Gu, Chaochen and Tang, Hao and Ye, Jieping},
  booktitle=NAACL,
  pages = "2289--2299",
  year={2025}
}

@INPROCEEDINGS{yin2025lifting,
author = {Yin, Hao and Si, Gunagzong and Wang, Zilei},
booktitle = CVPR,
title = {Lifting the Veil on Visual Information Flow in MLLMs: Unlocking Pathways to Faster Inference},
year = {2025},
pages = {9382-9391},
}

@InProceedings{wang2025knowledge,
    author    = {Wang, Sudong and Zhang, Yunjian and Zhu, Yao and Li, Jianing and Wang, Zizhe and Liu, Yanwei and Ji, Xiangyang},
    title     = {Towards Understanding How Knowledge Evolves in Large Vision-Language Models},
    booktitle = CVPR,
    year      = {2025},
    pages     = {29858-29868}
}

@InProceedings{liu2024seeing,
    author    = {Liu, Yexin and Liang, Zhengyang and Wang, Yueze and Wu, Xianfeng and Tang, Feilong and He, Muyang and Li, Jian and Liu, Zheng and Yang, Harry and Lim, Sernam and Zhao, Bo},
    title     = {Unveiling the Ignorance of MLLMs: Seeing Clearly, Answering Incorrectly},
    booktitle = CVPR,
    month     = {June},
    year      = {2025},
    pages     = {9087-9097}
}

@article{chen2023shikra,
  title={Shikra: Unleashing Multimodal LLM's Referential Dialogue Magic},
  author={Chen, Keqin and Zhang, Zhao and Zeng, Weili and Zhang, Richong and Zhu, Feng and Zhao, Rui},
  journal={arXiv preprint arXiv: 2306.15195},
  year={2023}
}

@article{chen2024internvl2,
      title={How Far Are We to GPT-4V? Closing the Gap to Commercial Multimodal Models with Open-Source Suites}, 
      author={Zhe Chen and Weiyun Wang and Hao Tian and Shenglong Ye and Zhangwei Gao and Erfei Cui and Wenwen Tong and Kongzhi Hu and Jiapeng Luo and Zheng Ma and Ji Ma and Jiaqi Wang and Xiaoyi Dong and Hang Yan and Hewei Guo and Conghui He and Botian Shi and Zhenjiang Jin and Chao Xu and Bin Wang and Xingjian Wei and Wei Li and Wenjian Zhang and Bo Zhang and Pinlong Cai and Licheng Wen and Xiangchao Yan and Min Dou and Lewei Lu and Xizhou Zhu and Tong Lu and Dahua Lin and Yu Qiao and Jifeng Dai and Wenhai Wang},
      year={2024},
      journal={arXiv preprint arXiv: 2404.16821},
}

@inproceedings{
  wang2025mllmseedynamiccorrection,
  title={{MLLM} can see? Dynamic Correction Decoding for Hallucination Mitigation},
  author={Chenxi Wang and Xiang Chen and Ningyu Zhang and Bozhong Tian and Haoming Xu and Shumin Deng and Huajun Chen},
  booktitle=ICLR,
  year={2025},
}

@inproceedings{li2023contrastivedecodingopenendedtext,
    title = "Contrastive Decoding: Open-ended Text Generation as Optimization",
    author = "Li, Xiang Lisa  and
      Holtzman, Ari  and
      Fried, Daniel  and
      Liang, Percy  and
      Eisner, Jason  and
      Hashimoto, Tatsunori  and
      Zettlemoyer, Luke  and
      Lewis, Mike",
    booktitle = "ACL",
    year = "2023",
    pages = "12286--12312",
}

@InProceedings{chen2024fastv,
author="Chen, Liang
and Zhao, Haozhe
and Liu, Tianyu
and Bai, Shuai
and Lin, Junyang
and Zhou, Chang
and Chang, Baobao",
title="An Image is Worth 1/2 Tokens After Layer 2: Plug-and-Play Inference Acceleration for Large Vision-Language Models",
booktitle=ECCV,
year="2025",
pages="19--35",
}
}

\end{document}